\pdfoutput=1

\documentclass[11pt]{article}

\usepackage[preprint]{acl}

\usepackage{times}
\usepackage{latexsym}
\usepackage{algorithm}
\usepackage{algpseudocode}
\usepackage{amsmath}
\usepackage{amsfonts}
\usepackage[T1]{fontenc}
\usepackage{array}
\newcolumntype{C}[1]{>{\centering\arraybackslash}m{#1}}

\usepackage[utf8]{inputenc}
\definecolor{citecolor}{HTML}{2980b9}
\definecolor{linkcolor}{HTML}{c0392b}
\usepackage{microtype}

\usepackage{xcolor}         
\usepackage{color, colortbl}
\definecolor{citecolor}{HTML}{2980b9}
\definecolor{linkcolor}{HTML}{c0392b}
\definecolor{darkorange}{HTML}{FF8C00}
\definecolor{chocolate}{HTML}{D2691E}
\definecolor{darkgreen}{HTML}{006400}
\definecolor{darkblue}{HTML}{00008B}
\definecolor{mediumblue}{HTML}{0000CD}
\definecolor{dodgerblue}{HTML}{1E90FF}
\definecolor{royalblue}{HTML}{4169E1}
\definecolor{shadecolor}{RGB}{237,237,237}
\definecolor{backred}{RGB}{255, 190, 190}
\definecolor{backblue}{RGB}{210, 230, 250}

\definecolor{zrrgreen}{HTML}{008000}
\definecolor{zrrblue}{HTML}{4682B4}
\definecolor{zrrred}{HTML}{B22222}
\usepackage{inconsolata}
\makeatletter
  \newcommand\figcaption{\def\@captype{figure}\caption}
  \newcommand\tabcaption{\def\@captype{table}\caption}
\makeatother

\definecolor{formalshade}{rgb}{0.95,0.95,1}

\definecolor{formalshade}{rgb}{0.95,0.95,1}
\newenvironment{formal}{%
  \MakeFramed{\advance\hsize-\width\FrameRestore}%
  \noindent\hspace{-4.55pt}%
  \begin{adjustwidth}{3pt}{3pt} 
  \vspace{-3pt}
}
{%
  \vspace{4pt}\end{adjustwidth}\endMakeFramed%
}

\usepackage{microtype}
\usepackage{graphicx}
\usepackage{amsmath}
\usepackage{amssymb}
\usepackage{booktabs} 
\usepackage{xcolor,colortbl}
\usepackage{amssymb}
\usepackage{subcaption}
\usepackage{makecell}
\usepackage{multirow}
\usepackage{framed}

\usepackage{enumitem}
\usepackage{wrapfig}
\usepackage{diagbox}
\usepackage{adjustbox}
\usepackage{tcolorbox}
\usepackage{changepage}
\newcommand\blfootnote[1]{%
  \begingroup
  \renewcommand\thefootnote{}\footnote{#1}%
  \addtocounter{footnote}{-1}%
  \endgroup
}
\definecolor{formalshade}{rgb}{0.95,0.95,1}

\title{
\textsc{SciVerse}:\vspace{0.1cm}\\Unveiling the Knowledge Comprehension and Visual Reasoning of\\LMMs on Multi-modal Scientific Problems
}
\author{Ziyu Guo$^*$ \quad Ray Zhang$^*$ \quad Hao Chen$^*$ \quad Jialin Gao$^*$\\ \textbf{Dongzhi Jiang} \quad \textbf{Jiaze Wang} \quad \textbf{Pheng-Ann Heng$^\dagger$} \\\\
        The Chinese University of Hong Kong\\
        \texttt{ziyuguo@link.cuhk.edu.hk}}

\begin{document}
\maketitle
\begin{abstract}
The rapid advancement of Large Multi-modal Models (LMMs) has enabled their application in scientific problem-solving, yet their fine-grained capabilities remain under-explored.
In this paper, we introduce \textsc{SciVerse}, a multi-modal scientific evaluation benchmark to thoroughly assess LMMs across 5,735 test instances in five distinct versions. We aim to investigate three key dimensions of LMMs: \textit{scientific knowledge comprehension}, \textit{multi-modal content interpretation}, and \textit{Chain-of-Thought (CoT) reasoning}. 
To unveil whether LMMs possess sufficient scientific expertise, we first transform each problem into three versions containing different levels of knowledge required for solving, i.e., Knowledge-free, -lite, and -rich. Then, to explore how LMMs interpret multi-modal scientific content, we annotate another two versions, i.e., Vision-rich and -only, marking more question information from texts to diagrams. Comparing the results of different versions, \textsc{SciVerse} systematically examines the professional knowledge stock and visual perception skills of LMMs in scientific domains. 
In addition, to rigorously assess CoT reasoning, we propose a new scientific CoT evaluation strategy, conducting a step-wise assessment on knowledge and logical errors in model outputs. Our extensive evaluation of different LMMs on \textsc{SciVerse} reveals critical limitations in their scientific proficiency and provides new insights into future developments.  Project page: \url{https://sciverse-cuhk.github.io}.
\end{abstract}
\blfootnote{\ This project is initially released in September, 2024}
\blfootnote{$^*$ Equal Contributions  \ $^\dagger$ Corresponding Author}

\section{Introduction}
In recent years, the rapid advancement of large models, i.e., Large Language Models (LLMs)~\cite{OpenAI2023ChatGPT, touvron2023llama, touvron2023llama2, vicuna2023} and Large Multi-modal Models (LMMs)~\cite{liu2023llava, openai2023gpt4v, zhang2024llamaadapter, gao2024sphinx}, has significantly expanded the frontiers of various modalities and scenarios, such as text~\cite{OpenAI2023GPT4TR,o1,guo2025deepseek}, 2D images~\cite{openai2023gpt4v,zhang2023llava,zhu2023minigpt,zhang2024llama}, and 3D point clouds~\cite{guo2024sam2point,xu2023pointllm,guo2023point,jia2024lift3d}. Notably, LMMs have demonstrated promising potential in addressing multi-modal scientific problems across diverse domains, including physics, chemistry, and biology.

Despite efforts to develop scientific datasets with visual content as evaluation benchmarks~\cite{lu2022learn,yue2023mmmu,yue2024mmmuprorobustmultidisciplinemultimodal}, existing approaches primarily assess LMMs through basic testing, where models directly solve original problems and are compared based on overall accuracy. However, we identify that effective problem-solving in this domain requires three key skills: \textit{scientific knowledge comprehension}, \textit{multi-modal content interpretation}, and \textit{Chain-of-Thought (CoT) reasoning}. Consequently, the fine-grained scientific capabilities of LMMs remain insufficiently explored, lacking a detailed and thorough examination within the research community.

\begin{figure*}[t!]
    \centering
    \vspace{2pt}
    \includegraphics[width=\linewidth]{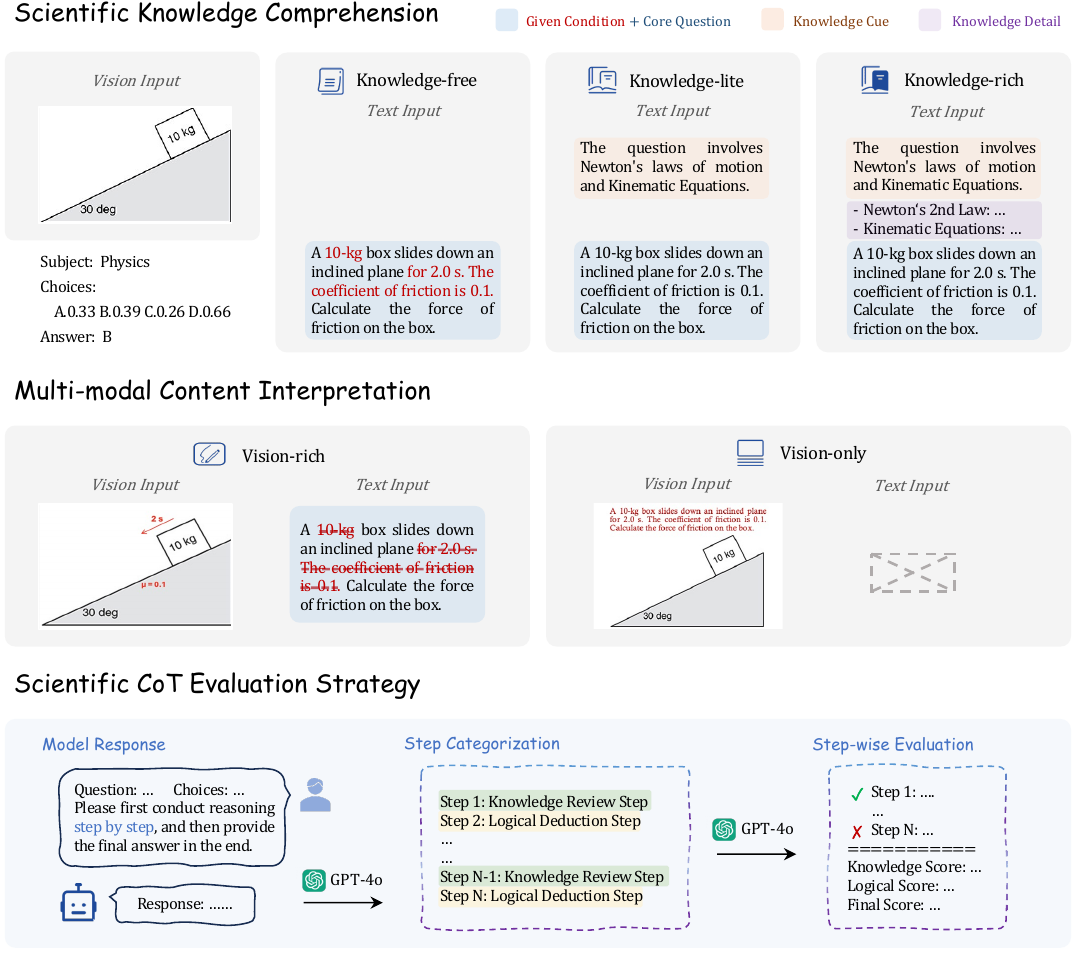}    \caption{\textbf{Overview of Five Problem Versions and our Scientific CoT Evaluation Strategy in \textsc{SciVerse.}} To unveil the scientific knowledge comprehension (Top), we first transform each problem into three versions integrating different levels of expertise knowledge. Then, to examine the multi-modal content interpretation (Middle), we further annotate two problem versions with varying vision-language information. We introduce a specialized scientific evaluation strategy (Bottom) to assess the fine-grained reasoning capabilities of LMMs.}
    \label{fig2}
\end{figure*}

In this paper, we introduce \textbf{\textsc{SciVerse}}, a comprehensive evaluation benchmark to assess LMMs on multi-modal scientific problems. Our curated dataset comprises 1,147 meticulously collected problems and 5,735 newly annotated test instances across five distinct versions, covering difficulty levels from high school to college. 
Specifically, to investigate the three key skills aforementioned, we aim to explore the following questions regarding scientific problem-solving as outlined in Figure~\ref{fig2}.

\begin{enumerate}
    \item \textit{\textbf{Do LMMs possess sufficient scientific knowledge to solve the problems?}}
    Unlike general visual scenarios, scientific problem-solving requires LMMs to have prior knowledge of specific subjects. Previous benchmarks do not differentiate between errors caused by a lack of knowledge and deficiencies in logical reasoning. To address this, we manually transform each problem of \textsc{SciVerse} into three versions with increasing levels of embedded knowledge within question texts: Knowledge-free, Knowledge-lite, and Knowledge-rich. By exposing LMMs to different depths of domain expertise, we systematically analyze how knowledge comprehension impacts scientific problem-solving.

    \item \textit{\textbf{Can LMMs effectively interpret question information from multi-modal content?}}
    In existing benchmarks, problem conditions are primarily presented in textual form, enabling LMMs to process them through language modeling. However, in real-world scenarios, key information is often embedded in diagrams, or even the entire question is printed as visual input (e.g., scanned documents, handwritten notes, or screenshots). Thus, it is essential to evaluate how LMMs perform when question content is progressively shifted from text to visual modalities. To this end, we further annotate the problems in \textsc{SciVerse} into two additional versions: Vision-rich and Vision-only. These versions systematically measure LMMs’ perception and OCR capabilities to retrieve and process multi-modal contexts in scientific problems.

    \item \textit{\textbf{Is CoT reasoning effective in improving the accuracy of solving scientific problems?}}
    Rather than directly providing a final answer, Chain-of-Thought (CoT) reasoning breaks the problem-solving process into a sequence of logical steps. In the context of scientific problems, the intermediate steps typically fall into two categories: knowledge review and logical deduction. Existing benchmarks generally assess CoT performance based on direct answer accuracy or a binary `True' or `False' metric. In contrast, we propose a new scientific CoT evaluation strategy using GPT-4o~\cite{openai2024gpt4o}. Our approach first extracts key steps from the model’s output and then performs a step-wise analysis, identifying both knowledge and reasoning errors. This methodology offers a more comprehensive evaluation of the CoT reasoning capabilities of LMMs.
    
\end{enumerate}

With five curated problem versions and a detailed CoT evaluation, our benchmark challenges LMMs to demonstrate not only expert knowledge but also their ability to integrate and reason across multiple modalities under varying levels of complexity. We evaluate a wide range of popular LMMs on \textsc{SciVerse}, offering unique insights to the research community. Our findings reveal that closed-source LMMs outperform open-source LMMs in both knowledge comprehension and visual perception in scientific domains. However, both categories of models struggle with Vision-only problems, which resemble real-world scenarios. Additionally, closed-source models exhibit stronger CoT reasoning capabilities, producing higher-quality reasoning steps.

Our contributions are threefold: 

\begin{itemize}
\item We present \textbf{\textsc{SciVerse}}, a multi-modal evaluation benchmark specifically designed to assess scientific reasoning across various disciplines. For the first time, \textsc{SciVerse} highlights three critical challenges that LMMs face in scientific problem-solving.

\item We develop a set of five problem versions that target distinct scientific reasoning challenges, addressing previous evaluation limitations in knowledge comprehension and multi-modal interpretation of LMMs.

\item We introduce a scientific CoT evaluation strategy, focusing on step-wise errors in both knowledge review and reasoning deduction. This approach offers a comprehensive analysis of LMMs' scientific CoT capabilities. \end{itemize}

\section{\textsc{SciVerse}}

In Section~\ref{s2.1}, we first present an overview of \textsc{SciVerse}, including dataset statistics and the collection process. Then, we respectively illustrate our methodology on the three critical aspects of assessing LMMs: scientific knowledge comprehension (Section~\ref{s2.2}), multi-modal content interpretation (Section~\ref{s2.3}), and Chain-of-Thought (CoT) reasoning evaluation (Section~\ref{s2.4}).

\subsection{Dataset Overview}
\label{s2.1}

To comprehensively evaluate scientific reasoning, we curate a diverse set of problems spanning a wide range of disciplines and knowledge domains.

\begin{figure*}[t]
\centering
\begin{minipage}[c]{0.39\textwidth}
\small
\centering
  \centering
  \begin{adjustbox}{width=\linewidth}
   \begin{tabular}{lr}
 \toprule
 \textbf{Statistic} & \textbf{Number} \\
 \midrule
  Total questions & 5,735 \\
  Questions of each version & 1,147 \\
  \midrule
  \rowcolor[gray]{.92}\textit{Knowledge-free} &\\
 Maximum question length & 1,353 \\
 Average question length & 254.3\\
 \rowcolor[gray]{.92}\textit{Knowledge-lite} &\\
 Maximum question length & 1,991 \\
 Average question length & 491.6\\
 \rowcolor[gray]{.92}\textit{Knowledge-rich} &\\
 Maximum question length & 2,768 \\
 Average question length & 842.2\\
  \rowcolor[gray]{.92}\textit{Vision-rich} &\\
 Maximum question length & 1,239 \\
 Average question length & 227.5\\
 \rowcolor[gray]{.92}\textit{Vision-only} &\\
 Maximum question length & 0 \\
 Average question length & 0\\
 \bottomrule
 \end{tabular}
 \end{adjustbox}
  \tabcaption{\textbf{Key Statistics of \textsc{SciVerse}.}}
  \label{tab1}
\end{minipage}
\qquad
\begin{minipage}[c]{0.49\textwidth}
\centering
\vspace{0.6cm}
\includegraphics[width=0.86\linewidth]{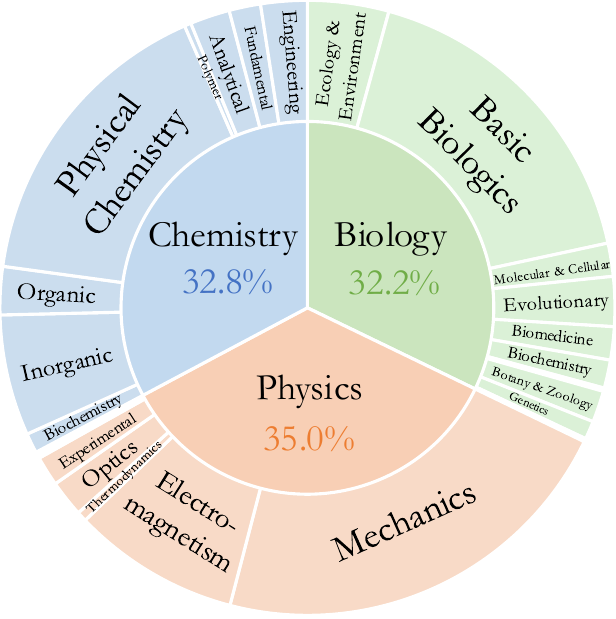}
\vspace{0.2cm}
\caption{\textbf{Subject Distribution of \textsc{SciVerse.}} The dataset contains 2,010 questions from Physics, 1,880 from Chemistry, and 1,845 from Biology.}
\vspace{0.2cm}
\label{fig3}
\end{minipage}
\end{figure*}

\paragraph{Data Statistics.}
Table~\ref{tab1} and Figure~\ref{fig3} provide an overview of the key statistics and subject distribution of \textsc{SciVerse}. The dataset consists of 5,735 problems, divided across three major domains: Physics, Chemistry, and Biology. These subjects are further broken down into 21 distinct scientific topics, allowing for an evaluation of problem-solving performance at a granular level. \textsc{SciVerse} includes five different problem versions, each consisting of 1,147 instances, designed to assess both the knowledge expertise and visual perception capabilities of LMMs. 
With more knowledge content integrated, the question length, from Knowledge-free, Knowledge-lite, to Knowledge-rich versions, also increases. As the information is gradually transited from texts to diagrams, the question length decreases from Knowledge-lite, Vision-rich, to Vision-only versions.

\begin{figure*}[t!]
    \centering
    \includegraphics[width=\linewidth]{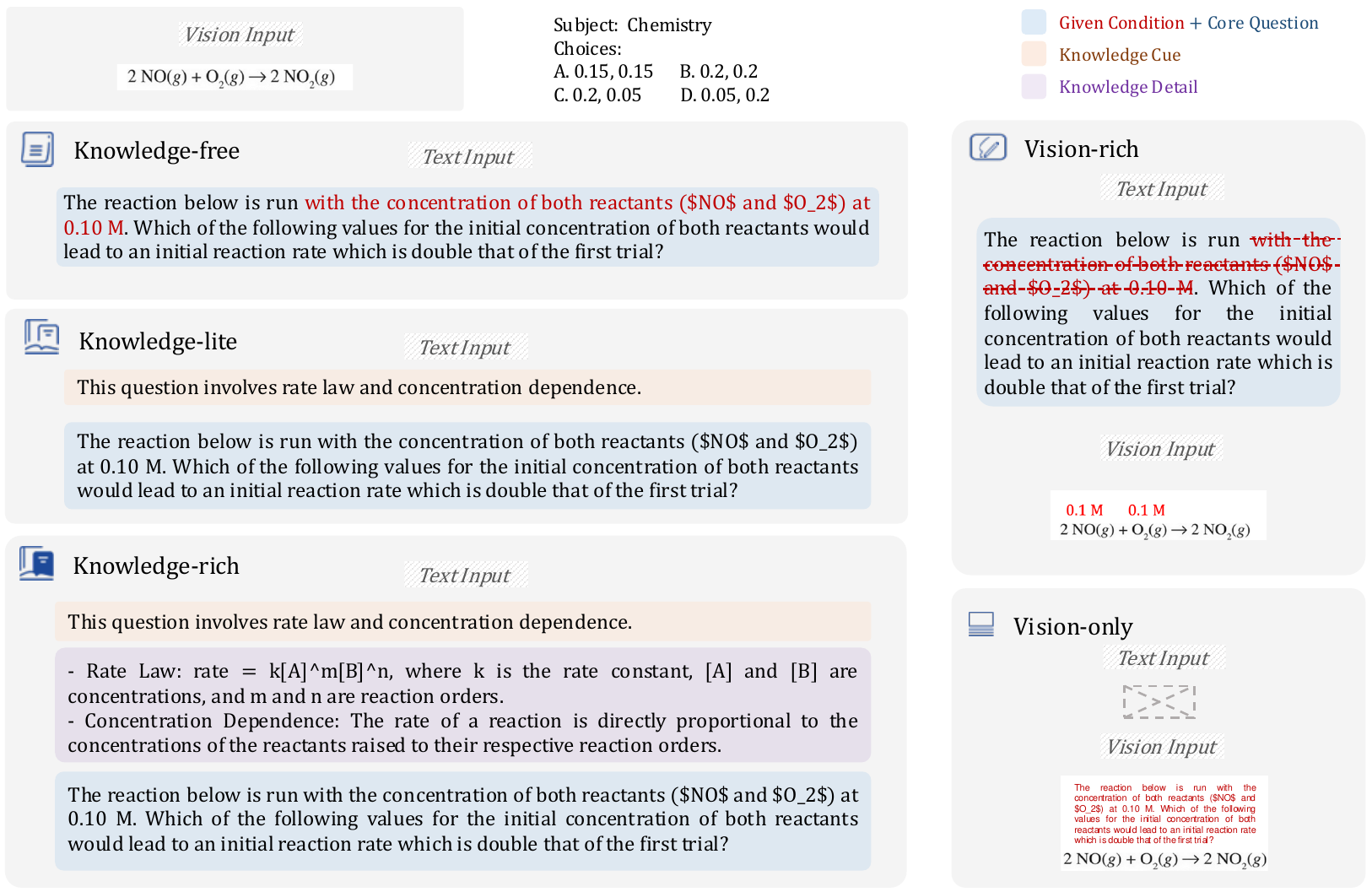}    
    \caption{\textbf{Examples of Five Problem Versions in \textsc{SciVerse}.} For each problem in \textsc{SciVerse}, we first create the Knowledge-free version by removing all knowledge content from the question text. Next, we add knowledge cues and details to produce the Knowledge-lite and Knowledge-rich versions. Additionally, starting from the Knowledge-free version, we generate two more versions, Vision-rich and Vision-only, where the given condition and, ultimately, the entire question are transferred to the visual diagram.}
    \label{fig33}
\end{figure*}

\paragraph{Data Curation.}
To guarantee a comprehensive scope, we begin by reviewing publicly available scientific datasets, from which we curate an initial set of 1,200 problems sourced from three datasets: SceMQA~\cite{liang2024scemqa}, MMMU~\cite{yue2023mmmu}, and CMMMU~\cite{zhang2024cmmmu}. To maintain high quality, we engage eight PhD-level science experts to carefully evaluate and select problems based on the knowledge complexity and visual richness of problems. Subsequently, we translate all texts into English in a Latex format, and convert the problem types into multiple-choice questions. After a thorough review process, 1,147 problems are retained, each of which is then transformed into five different versions, as outlined in the following sections and illustrated in Figure~\ref{fig33}.

\subsection{Scientific Knowledge Comprehension}
\label{s2.2}

A key challenge for LMMs in solving scientific problems is their capability to comprehend sufficient domain knowledge, which is essential for understanding the question and performing multi-modal reasoning. To evaluate this, we manually transform each problem in \textsc{SciVerse} into three versions, each incorporating varying levels of scientific knowledge. By comparing the performance of an LMM across these three versions, we aim to investigate the impact of knowledge comprehension on scientific problem-solving.

\paragraph{Knowledge-free Version.}
We first eliminate all background knowledge from the question text, leaving only the core question, which includes the given condition (e.g., \textit{slides down for 2.0 s''}) and core question (e.g., \textit{calculate the force''}). This version presents a significant challenge for LMMs, as they must first interpret the question accurately and then relate it to the appropriate scientific knowledge for problem-solving. The content in both text and visual modalities is structured as follows:

{\small{Text Input: Given Condition + Core Question}}

{\small{Vision Input: Diagram}}

\paragraph{Knowledge-lite Version.}
Based on the previous version, we introduce a simple knowledge cue in the question text, indicating the high-level knowledge required for solving the problem. Typically, we provide related theorem names or formulation references at the beginning of the question, such as \textit{Newton's laws of motion''} or \textit{\textit{Kinematic Equations}''}. These cues help guide the LMMs in interpreting the problem and allow us to assess whether their performance improves when provided with basic background knowledge, compared to Knowledge-free results. The content is structured as:

{\small{Text Input: Knowledge Cue + Given Condition + Core Question}}

{\small{Vision Input: Diagram}}

\paragraph{Knowledge-rich Version.}
In this version, we further enrich the problem with detailed scientific information, such as specific equations and the application method of a relevant theorem (e.g., \textit{``This law states that the net force (F) acting on an object is equal to the product of its mass (m) and its acceleration (a).''}). By comparing performance in the Knowledge-rich and -lite versions, we can determine whether LMMs truly comprehend the expertise required and whether their performance improves when provided with more detailed background information. The content is structured as:

{\small{Text Input: Knowledge Cue + Knowledge Detail + Given Condition + Core Question}}

{\small{Vision Input: Diagram}}

\subsection{Multi-modal Content Interpretation}
\label{s2.3}

In contrast to LLMs, LMMs must accurately interpret the diagram input and integrate visual information with the textual question for effective question-answering. Therefore, we focus on evaluating the visual perception capabilities of LMMs in the context of scientific diagrams. To this end, we transform each problem in \textsc{SciVerse} into two additional versions that progressively shift the balance of question information from text to diagrams, which are more similar to real-world scenarios.

\begin{figure*}[t!]
    \centering
    \includegraphics[width=\linewidth]{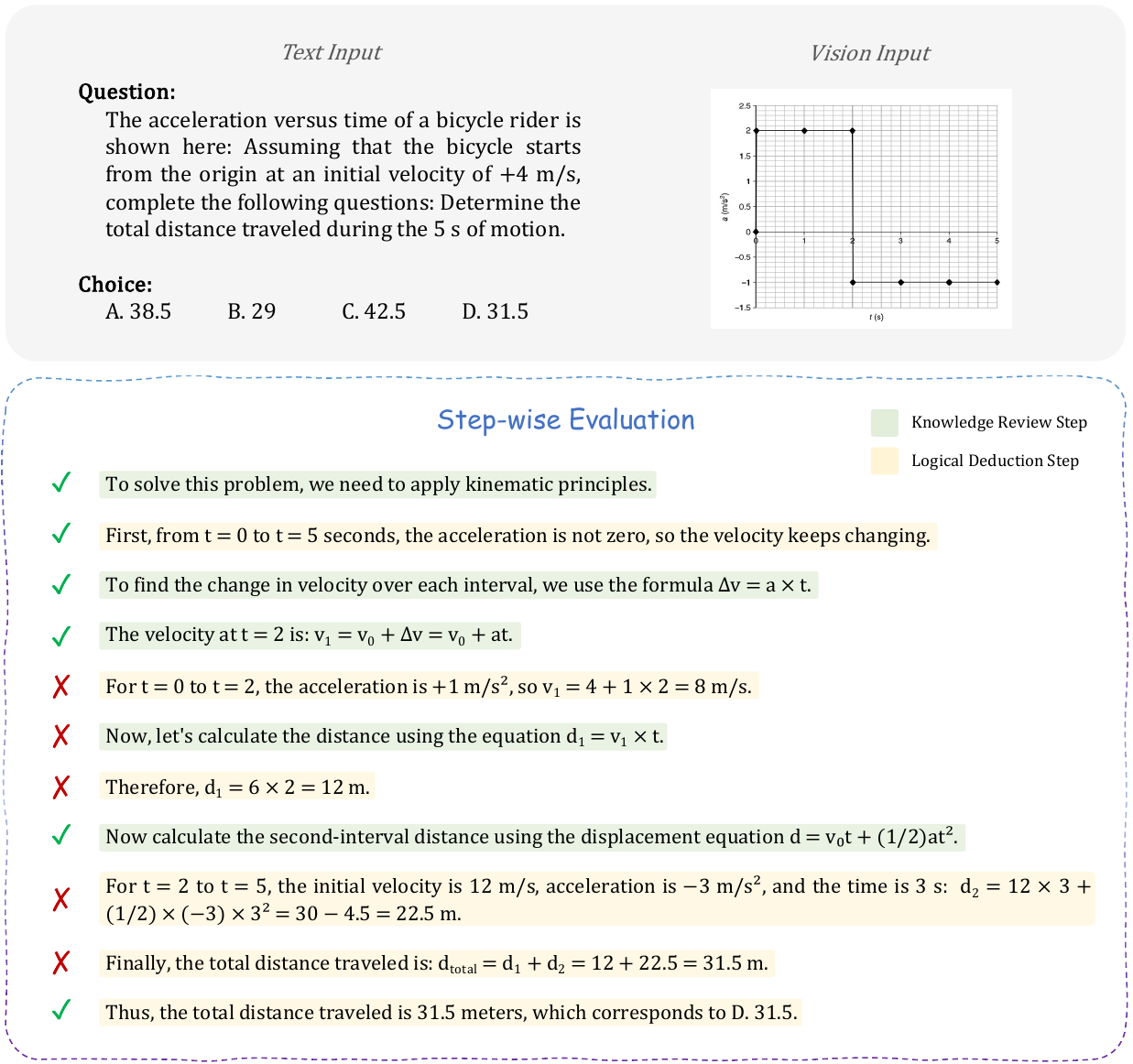}
    \caption{\textbf{Examples of the Scientific CoT Evaluation Strategy.} For reasoning responses from LMMs, we prompt GPT-4o~\cite{openai2024gpt4o} to perform two evaluation stages, i.e., step categorization and step-wise evaluation. We categorize the intermediate steps into two types: knowledge review and logical reasoning.}
    \label{fig4}
\end{figure*}

\paragraph{Vision-rich Version.}
On top of the Knowledge-lite version, we remove most of the problem conditions from the question text (e.g., \textit{``slides down for 2.0 s''}) and instead annotate them directly onto the diagram, provided they can be suitably represented visually. This version challenges LMMs to rely more on the visual modality for extracting critical information, reducing the reliance on textual content and testing their true multi-modal interpretation capabilities in scientific problem-solving. The content is structured as follows:

{\small{Text Input: Core Question}}

{\small{Vision Input: Diagram + Given Condition}}

\paragraph{Vision-only Version.}
In this version, we take the integration of visual information a step further by embedding the entire question directly onto the diagram, eliminating any textual input. This setup closely mirrors real-world scenarios where users capture an image or screenshot of a problem. Without any textual cues, vision-only problems present the most challenging evaluation for LMMs, which assess their capabilities in knowledge comprehension, OCR, and visual perception. The content is structured as follows:

{\small{Text Input: None}}

{\small{Vision Input: Diagram + Given Condition + Core Question}}

\subsection{Scientific CoT Evaluation Strategy}
\label{s2.4}
For complex scenarios, utilizing CoT~\cite{wei2022chain} to perform step-by-step reasoning is essential for improving the problem-solving accuracy of LMMs. While some previous scientific benchmarks~\cite{yue2024mmmuprorobustmultidisciplinemultimodal} have reported the CoT performance, they still rely on a binary `True' or `False' metric based on the final answer, overlooking the quality of the intermediate steps during reasoning. To address this gap, we propose a specialized scientific CoT evaluation strategy designed to assess the fine-grained CoT capabilities of LMMs in scientific problem-solving. This strategy involves two sequential stages using GPT-4o~\cite{openai2024gpt4o} as shown in Figure~\ref{fig4}.

\paragraph{Step Categorization.}
For model responses generated using CoT prompting~\cite{kojima2022large}, we first apply GPT-4o to extract the key steps from the extended reasoning sequence and categorize them into two types:

\begin{itemize} 
    \item \textbf{Knowledge Review Step} refers to the process of quoting or recalling relevant expert knowledge during problem-solving (e.g., \textit{``we need to apply kinematic principles''}). These review steps assist LMMs in subsequent reasoning but may be prone to errors, such as quoting an irrelevant theorem or misrecalling equations.
    
    \item \textbf{Logical Deduction Step} involves applying logical reasoning to derive an intermediate or final conclusion, which can be either a calculated result (e.g., \textit{``d$_1$ = 6 × 2 = 12 m''}) or a knowledge-based inference (e.g., \textit{``so the velocity keeps changing''}).
    This step may encounter errors, such as incorrect calculations, improper substitutions, or flawed inferences.
    
\end{itemize}

\begin{table*}[!t]
\small
\centering
\begin{adjustbox}{width=\linewidth}
    \begin{tabular}{l|@{\extracolsep{6pt}}C{0.8cm}@{\extracolsep{6pt}}C{0.8cm}|@{\extracolsep{6pt}}C{0.8cm}@{\extracolsep{6pt}}C{0.8cm}@{\extracolsep{6pt}}C{0.8cm}@{\extracolsep{6pt}}C{0.8cm}|@{\extracolsep{6pt}}C{0.8cm}@{\extracolsep{6pt}}C{0.8cm}@{\extracolsep{6pt}}C{0.8cm}@{\extracolsep{6pt}}C{0.8cm}|@{\extracolsep{6pt}}C{0.8cm}@{\extracolsep{6pt}}C{0.8cm}@{\extracolsep{6pt}}C{0.8cm}@{\extracolsep{6pt}}C{0.8cm}|@{\extracolsep{6pt}}C{0.8cm}@{\extracolsep{6pt}}C{0.8cm}@{\extracolsep{6pt}}C{0.8cm}@{\extracolsep{6pt}}C{0.8cm}|@{\extracolsep{6pt}}C{0.8cm}@{\extracolsep{6pt}}C{0.8cm}@{\extracolsep{6pt}}C{0.8cm}@{\extracolsep{6pt}}C{0.8cm}}
    \toprule
    \multirow{2}*{\makecell*[l]{\large Model}}    &\multicolumn{2}{c|}{\makecell*[c]{\bf All}}
    &\multicolumn{4}{c|}{Knowledge-rich} 
    &\multicolumn{4}{c|}{Knowledge-free}
    &\multicolumn{4}{c|}{Knowledge-lite}
    &\multicolumn{4}{c|}{Vision-rich}
    &\multicolumn{4}{c}{Vision-only}\\    
    \cmidrule{2-23}
    & \bf Acc & \bf Sci-CoT  & Acc &  Sci-CoT & Sci-CoT$_K$ &  Sci-CoT$_L$& Acc & Sci-CoT & Sci-CoT$_K$ & Sci-CoT$_L$& Acc & Sci-CoT & Sci-CoT$_K$ & Sci-CoT$_L$& Acc & Sci-CoT & Sci-CoT$_K$ & Sci-CoT$_L$& Acc & Sci-CoT & Sci-CoT$_K$ & Sci-CoT$_L$\\
    \midrule
    \rowcolor{gray!10}\multicolumn{23}{c}{\textit{Baseline}}\\
    \cmidrule{1-23}
    Random Chance & \bf 22.7 & \bf - & 22.7 & -  & -  & - & 22.7 & -  & -  & - & 22.7 & -  & -  & - & 22.7 & -  & -  & - & 22.7 & -  & -  & - \\
    \midrule
    \rowcolor{gray!10}\multicolumn{23}{c}{\textit{Closed-source LMMs}}\\
    \cmidrule{1-23}
    GPT-4V &\bf 45.7 &\bf 52.3 &47.1 &55.8 &72.3 &39.3 &46.8 &54.1& 69.3&41.8 &46.6 &52.9& 66.4&39.4 &46.0 &52.0&65.2&38.8 &42.1&50.7&60.4 &41.0\\ 
    Gemini-1.5-Pro &\bf 49.5 &\bf 58.6 &50.8 &62.2&78.2 &46.2 &50.7 &60.9&76.3&45.5 &50.5 &58.4&70.7&46.1 &49.9 &57.3&68.4&46.2 &45.9&55.2&64.3 &46.1\\
    Claude-3.5-Sonnet &\bf 52.8 &\bf 62.4 &54.1 &66.9&80.2&53.6 &53.9 &63.4&78.8&48.0&53.7 &62.5&75.3&49.7 &53.1 &61.3&72.5&50.1 &49.3 &59.3&69.9 &48.7 \\
    GPT-4o   &\bf \colorbox{backred!50}{54.0} &\bf \colorbox{backred!50}{66.7} &\colorbox{backred!50}{55.3} &\colorbox{backred!50}{70.8}&84.6&57.0 &\colorbox{backred!50}{55.2} &\colorbox{backred!50}{67.8}&80.3&55.3 &\colorbox{backred!50}{55.0} &\colorbox{backred!50}{66.4}&78.2&54.6 &\colorbox{backred!50}{54.4} &\colorbox{backred!50}{66.4}&76.3&56.5 &\colorbox{backred!50}{50.2} &\colorbox{backred!50}{64.0} &71.4 &56.6\\
    \midrule
    \rowcolor{gray!10}\multicolumn{23}{c}{\textit{Open-source LMMs}}\\
    \cmidrule{1-23}
    SPHINX-Tiny (1.1B) &\bf 27.6&\bf 30.2 &28.9	&34.7&38.2 &31.2&29.1	&31.4&34.4&28.4&26.7	&29.8&31.4&28.2&26.1	&29.5&30.9&28.1&27.2&26.5&28.2 &24.8\\
     MiniGPT-v2 (7B) &\bf 30.0 &\bf 34.1 &30.5&37.8&41.2&34.4&29.6	&35.7&38.2&33.2&31.5&39.9&34.0&45.8&31.6	&32.8&33.1&32.5&26.9 &30.3&30.9 &20.7\\
     ShareGPT4V (13B) &\bf 33.4 &\bf 36.9 &36.3	&41.3&44.6&38.0&34.7	&37.2&41.3&33.1&32.3	&36.9&38.7&35.1&32.3	&36.7&37.1&36.3&31.5 &32.5&33.9 &31.1\\
     LLaVA-1.5 (13B) &\bf 33.7 &\bf 38.2 &35.4	&42.8&45.1&40.5&35.0	&39.5&41.9&37.1&32.7	&38.2&39.7&36.7&33.0	&37.9&39.2&36.6&32.3 &34.0&34.9 &33.1 \\
     LLaVA-NeXT (8B) &\bf 36.4&\bf 39.4 &39.0	&43.1&48.2&38.0&39.1	&40.6&46.8&34.4&36.6	&39.0&43.1&34.9&36.1 &39.3&42.3&36.3&31.3 &35.6&38.2 &33.0\\
    InternLM-XC2. (7B) &\bf 36.7&\bf 40.9 &39.9	&44.4&49.3&39.5&38.8	&41.8&46.3&37.3&37.6	&40.6&44.2&37.0&35.5	&40.6&43.0&38.2&31.6 &37.8&40.2&55.6\\ 
    SPHINX-MoE (8$\times$7B)  &\bf 37.3&\bf 41.1 &41.3	&44.0&48.7&39.3&38.8	&41.9&47.2&36.6&38.9	&39.9&43.7&36.1&36.3	&41.3&44.2&38.4&31.4&38.2&41.1&35.3\\
    SPHINX-Plus (13B) &\bf 37.3&\bf 41.2 &41.4	&44.5&49.1&39.9&38.7	&42.0&47.6&36.4&39.2	&41.0&44.9&37.1&37.4	&41.1&44.8&37.4&29.6&38.4&41.0&35.8\\
    InternVL-1.5 (26B) &\bf 39.0&\bf 46.3 &40.4	 &49.8&51.3&48.3&41.7	&47.3&49.9&44.7&39.5	&46.2&49.2&43.2&39.2	&46.1&48.8&43.4&34.3&42.5&41.3&43.7\\
    InternVL-2 (8B) &\bf 42.6 &\bf 49.9 &43.9 &53.8&59.2 &48.4 &43.7 &50.7&58.3&43.1 &43.2 &49.7&54.3&45.1 &42.9 &49.7&52.9&46.5 &39.1&46.1&49.2&43.0\\
    Qwen2-VL (7B) &\bf 44.7 &\bf \colorbox{backblue!75}{53.2} &46.2 &\colorbox{backblue!75}{57.3}&63.1&51.5 &45.9 &54.0&63.2&44.8 &45.8 &\colorbox{backblue!75}{53.3}&61.3&45.3 &45.0 &\colorbox{backblue!75}{53.1}&61.7&44.5 &40.4&\colorbox{backblue!75}{48.6}&60.0&37.2\\
    LLaVA-OneVision (7B) &\bf \colorbox{backblue!75}{46.1} &\bf 51.3 &\colorbox{backblue!75}{47.6} &54.6&61.7&47.5 &\colorbox{backblue!75}{47.2} &\colorbox{backblue!75}{54.0}&61.4&46.6 &\colorbox{backblue!75}{47.0} &51.1&60.3&41.9 &\colorbox{backblue!75}{46.5} &49.7&59.9&39.5 &\colorbox{backblue!75}{41.9} &47.3&59.7 &34.9\\
    \bottomrule
    \end{tabular}
\end{adjustbox}
\caption{\textbf{Evaluation Results on Five Problem Versions of \textsc{SciVerse}.} 
The `All' scores represent the average results across all five problem versions. The metric `Acc' refers to the binary `True' or `False' evaluation based solely on the final answer. `Sci-CoT' refers to our proposed scientific CoT evaluation strategy, averaging the scores of knowledge review and logical reasoning, denoted as `Sci-CoT$_K$' and `Sci-CoT$_L$' The highest scores for \colorbox{backred!50}{closed-source} and \colorbox{backblue!75}{open-source} LMMs are marked in red and blue, respectively.} 
\label{tab2}
\end{table*}

\paragraph{Step-wise Evaluation.}
Following the step categorization, we prompt GPT-4o to provide a fine-grained `True' or `False' judgment for each individual step. This step-wise evaluation thoroughly considers each intermediate step, offering insights into the detailed CoT reasoning capabilities of LMMs. Subsequently, we compute two average scores: one for the knowledge comprehension steps and another for the logical deduction steps. In contrast to the previous binary accuracy, our strategy, which generates two distinct scores, provides a more comprehensive assessment of the model's understanding of scientific knowledge and its proficiency in CoT reasoning.

\section{Experiment}

In Section~\ref{s3.1}, we first introduce our experimental settings, including the evaluation LMMs and implementation details. Then, in Section~\ref{s3.2}, we provide the performance comparison and insightful analysis on \textsc{SciVerse}.

\subsection{Evaluation Settings}
\label{s3.1}
\paragraph{Evaluation Models.}
We comprehensively assess a wide range of open-source and closed-source LMMs on \textsc{SciVerse}. Closed-source models include Gemini-1.5-Pro~\cite{team2023gemini}, Claude-3.5-Sonnet~\cite{anthropic2024claude35}, GPT-4V~\cite{openai2023gpt4v}, and GPT-4o~\cite{openai2024gpt4o}. Open-source models include MiniGPT-v2~\cite{chen2023minigpt}, LLaVA-1.5~\cite{liu2023improvedllava}, LLaVA-NeXT~\cite{liu2024llavanext}, LLaVA-OneVision~\cite{li2024llava-ov}, ShareGPT4V~\cite{Chen2023ShareGPT4VIL}, SPHINX series~\cite{gao2024sphinx}, InternLM-XComposer-2~\cite{dong2024internlm}, InternVL-1.5~\cite{chen2024far}, InternVL-2~\cite{chen2024far}, Qwen2-VL~\cite{Qwen2-VL}, and Qwen2.5-VL~\cite{qwen2.5-VL}.

\paragraph{Implementation Details.}
We adopt two metrics for evaluation. The first is the previous binary metric solely based on the final answer, termed \textbf{`Acc'}. We adopt an input prompt, \textit{``directly provide the answer''}, to guide LMMs to provide the final answer directly. The second is our proposed scientific CoT evaluation strategy. We term the scores of knowledge and logical errors as \textbf{`Sci-CoT$_K$'} and \textbf{`Sci-CoT$_L$'}, respectively, and denote their average score as \textbf{`Sci-CoT'}. We adopt an input CoT prompt, \textit{``perform reasoning step-by-step''}, to elicit step-wise reasoning output.
We evaluate all LMMs in a zero-shot setting without few-shot examples.
We also provide a baseline representing random chance by randomly selecting an option. All evaluation is conducted on NVIDIA A100 GPUs.

\subsection{Discussion and Analysis}
\label{s3.2}

In Table~\ref{tab2}, we present the detailed evaluation results of \textsc{SciVerse}. Based on the performance comparison, we derive several key observations:
\begin{itemize}
    \item \textit{As more knowledge is provided, open-source LMMs show greater improvement, whereas closed-source LMMs exhibit relatively smaller gains.}
    As we move from the Knowledge-free, Knowledge-lite, to Knowledge-rich versions, most LMMs demonstrate performance improvements as more knowledge cues and details are added to the question. Among these, closed-source LMMs, such as GPT-4o and Claude-3.5-Sonnet, display relatively stable results across all three versions. This stability suggests that these models inherently possess a greater depth of expertise knowledge and are better able to effectively leverage it for problem-solving. This trend is further supported by the results of `Sci-CoT$_K$', where closed-source models achieve higher accuracy in knowledge review compared to their open-source counterparts.

    \item \textit{When more information is shifted to vision input, open-source LMMs experience a significantly larger performance drop compared to closed-source LMMs.}
    From Knowledge-free to Vision-rich versions, most LMMs exhibit a noticeable performance decline. This suggests that, relative to text-based question information, LMMs face greater challenges when problem conditions are given as visual information. Such results highlight the limitations in the visual encoding quality and cross-modal understanding of current LMMs when applied to scientific diagrams. Additionally, closed-source LMMs show a smaller performance drop between the two problem versions, indicating their relatively stronger capabilities in scientific visual perception.

    \item \textit{The most challenging scenario for LMMs occurs with Vision-only problems, where all question information is embedded in diagrams.}
    The largest performance drop is observed between the Vision-rich and Vision-only versions for both closed-source and open-source LMMs. This indicates that LMMs struggle with low capabilities for the OCR and interpretation of question information embedded visually in diagrams. 
    Such a lack of reliable OCR capabilities and cross-modal integration severely hinders LMMs' potential to tackle scientific problems in real-world scenarios.

    \item \textit{Closed-source LMMs demonstrate notably stronger CoT reasoning capabilities than open-source LMMs.}
    When comparing the `Acc' and `Sci-CoT' scores across all problem versions, we observe a significant gap, with the CoT evaluation score being higher than the binary accuracy. This suggests that many intermediate steps may be correct, even when the final answer is incorrect. Such cases would be overlooked by the traditional binary accuracy metric, but our scientific CoT evaluation strategy effectively identifies and incorporates them into the final scores. Furthermore, the gap between the two scores is more pronounced in closed-source LMMs, indicating that closed-source models excel at CoT reasoning, producing higher-quality intermediate steps and more robust overall performance.
    
\end{itemize}

\section{Related Work}
\subsection{Multi-modal Scientific Benchmark}
Recent advances in LMMs have sparked significant interest in their mathematic~\cite{zhang2024mathverse, zhang2024mavis} and scientific reasoning capabilities, particularly in tasks involving visual interpretation. A spectrum of scientific benchmarks has emerged across different educational levels: ScienceQA~\cite{lu2022learn} targets elementary and secondary education, focusing on foundational scientific concepts. Moving to higher education, SceMQA~\cite{liang2024scemqa} introduces a comprehensive benchmark at the college entrance level, encompassing Mathematics, Physics, Chemistry, and Biology. At the collegiate level, MMMU~\cite{yue2023mmmu} and its enhanced version MMMU-Pro~\cite{yue2024mmmuprorobustmultidisciplinemultimodal} have emerged as broader benchmarks, spanning diverse fields from arts to technology. The multilingual expansion is demonstrated by CMMMU~\cite{zhang2024cmmmu}, which extends the evaluation framework to Chinese contexts. For advanced evaluation, OlympiadBench~\cite{he2024olympiadbench} incorporates challenging Mathematics and Physics Olympiad problems, testing LMMs' capabilities in solving exceptionally difficult problems. Meanwhile, some recent works~\cite{guo2025can, jiang2025mme, jiang2024mmsearch} also focus on the exploration of the \textit{Retrieval-Augmented Generation (RAG)} ability and \textit{Chain-of-Thought (CoT) reasoning} reasoning ability of the LMMs. 
Different from all previous works, our \textsc{SciVerse}, for the first time, investigate three critical issues within LMMs in scientific problem-solving, i.e., \textit{scientific knowledge comprehension}, \textit{multi-modal content interpretation}, and \textit{Chain-of-Thought (CoT) reasoning}, offering unique insights to the community.

\subsection{Large Multi-modal Models (LMMs)}
Recent advances in multi-modal AI have been marked by significant developments in LMMs, which combine the capabilities of LLMs and vision models to process diverse visual inputs. While proprietary models like GPT-4V~\cite{openai2023gpt4v}, Claude~\cite{anthropic2024claude35}, Gemini~\cite{team2023gemini}, and GPT-4o~\cite{openai2024gpt4o} have shown remarkable visual reasoning abilities, their closed nature has spurred the development of open-source alternatives. Early open-source LMMs like LLaVA~\cite{liu2023llava} and MiniGPT-4~\cite{zhu2023minigpt} paired CLIP-based image encoders~\cite{radford2021learning} with LLMs for multi-modal instruction tuning. Later models such as LLaVA-NeXT~\cite{li2024llava}, LLaVA-OneVision~\cite{li2024llava-ov}, ShareGPT4V~\cite{Chen2023ShareGPT4VIL}, InternVL~\cite{chen2024internvl}, SPHINX~\cite{lin2023sphinx}, and Qwen-VL~\cite{Qwen2-VL} expanded these capabilities through broader training datasets and advanced training strategies. In this paper, we aim to comprehensively evaluate their fine-grained capabilities in scientific domains, guiding the future developments of LMMs.

\section{Conclusion}
In this paper, we introduce \textsc{SciVerse}, a comprehensive multi-modal benchmark designed to evaluate the fine-grained capabilities of LMMs in scientific problem-solving. By transforming problems into multiple versions that vary in knowledge and modality, we investigate three critical dimensions of LMMs: scientific knowledge comprehension, multi-modal content interpretation, and CoT reasoning. Furthermore, our proposed scientific CoT evaluation strategy provides a deeper understanding of how LMMs handle knowledge and logical errors during problem-solving. The findings from our extensive evaluation of current state-of-the-art LMMs underscore the need for further advancements in their scientific proficiency and multi-modal reasoning capabilities. Moving forward, we hope \textsc{SciVerse} may serve as a foundation for future developments of LMMs in scientific fields.

\section*{Limitations}
Although our primary focus is on investigating the three critical issues of LMMs in scientific domains, rather than the breadth of evaluation, future work could expand \textsc{SciVerse} to include additional disciplines and scenarios, such as art, business, medicine, and social sciences.




\bibliography{custom}

\clearpage
\appendix

\section{GPT-4o Prompt for the CoT Scientific Evaluation Strategy}

\subsection{Step Categorization}
\begin{formal}
\textit{You will be provided with a step-by-step solution to a problem. Your task is to:  \\
1. Break the solution into the smallest possible steps, ensuring each step represents a single action or piece of reasoning.  \\
2. Classify each step as either:  \\
   \quad - \{K\}: \textbf{Knowledge Review Step} (facts, definitions, or prior knowledge used in the step).  \\
   \quad - \{L\}: \textbf{Logical Deduction Step} (deductions, calculations, or inferences made in the step). }
\end{formal}

\subsection{Step-wise Evaluation}
\begin{formal}
\textit{You will be provided with a list of steps from a solution for a scientific question, each classified as either Knowledge Review Step (\{K\}) or Logical Deduction Step (\{L\}). Your task is to assign a correctness score to each step:  \\
   - \{1\}: \textbf{Correct} (the knowledge is relevant, sufficient, and accurate, or the reasoning is logically valid).\\  
   - \{0\}: \textbf{Incorrect} (the knowledge is irrelevant, insufficient, or inaccurate, or the reasoning is flawed).  }
\end{formal}

\vspace{0.4cm}
\section{Additional Examples}
\label{sec:appendix}

\begin{figure*}[t!]
    \centering
    \includegraphics[width=\linewidth]{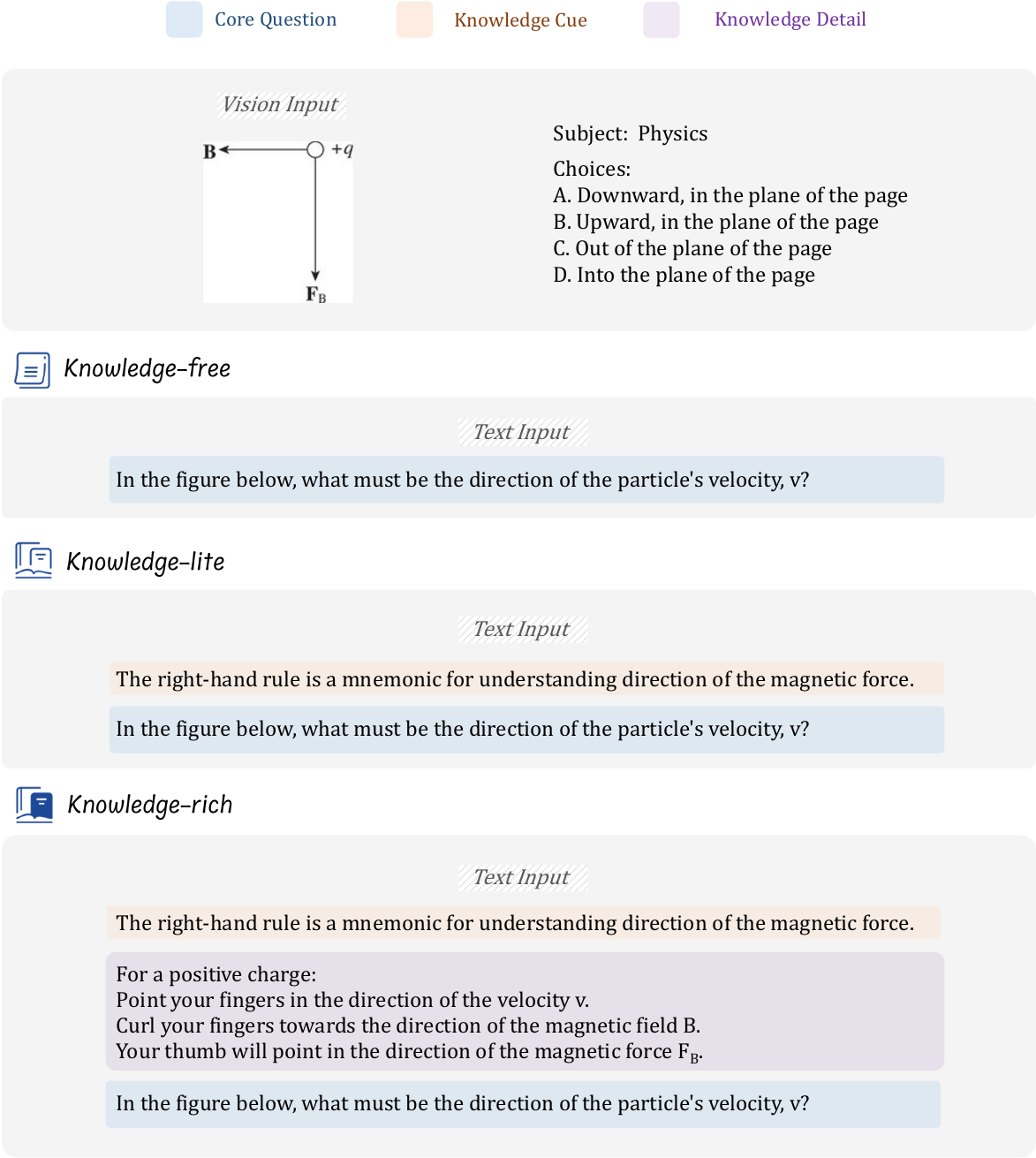}
    \caption{\textbf{Examples of Different Problem Versions in \textsc{SciVerse}.}}
    \label{apfig1}
\end{figure*}

In Figures~\ref{apfig1}$\sim$\ref{apfig9}, we provide more examples of different problem versions in \textsc{SciVerse}.

\begin{figure*}[t!]
    \centering
    \includegraphics[width=\linewidth]{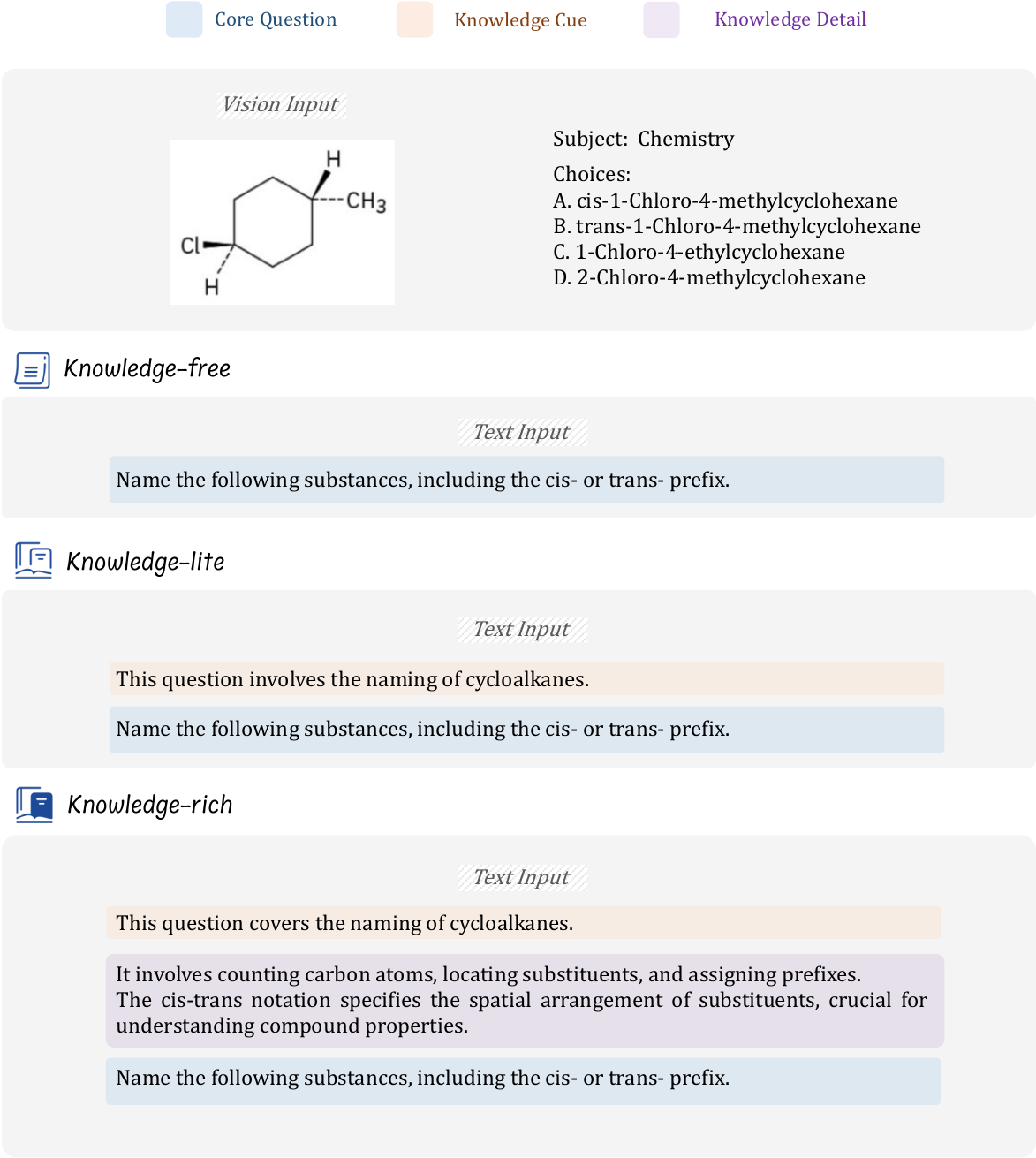}
    \caption{\textbf{Examples of Different Problem Versions in \textsc{SciVerse}.}}
    \label{apfig2}
\end{figure*}

\begin{figure*}[t!]
    \centering
    \includegraphics[width=\linewidth]{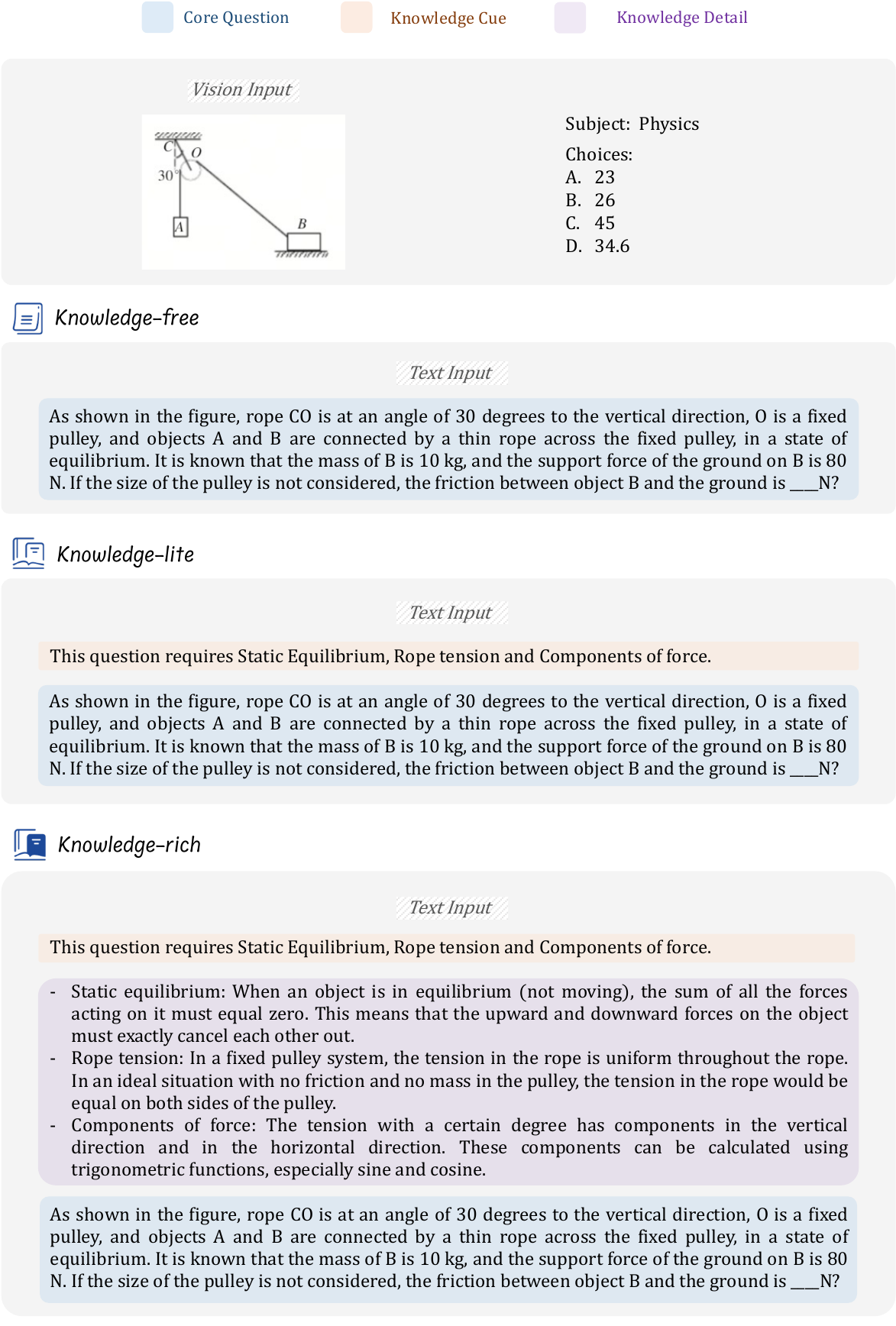}
    \caption{\textbf{Examples of Different Problem Versions in \textsc{SciVerse}.}}
    \label{apfig3}
\end{figure*}

\begin{figure*}[t!]
    \centering
    \includegraphics[width=\linewidth]{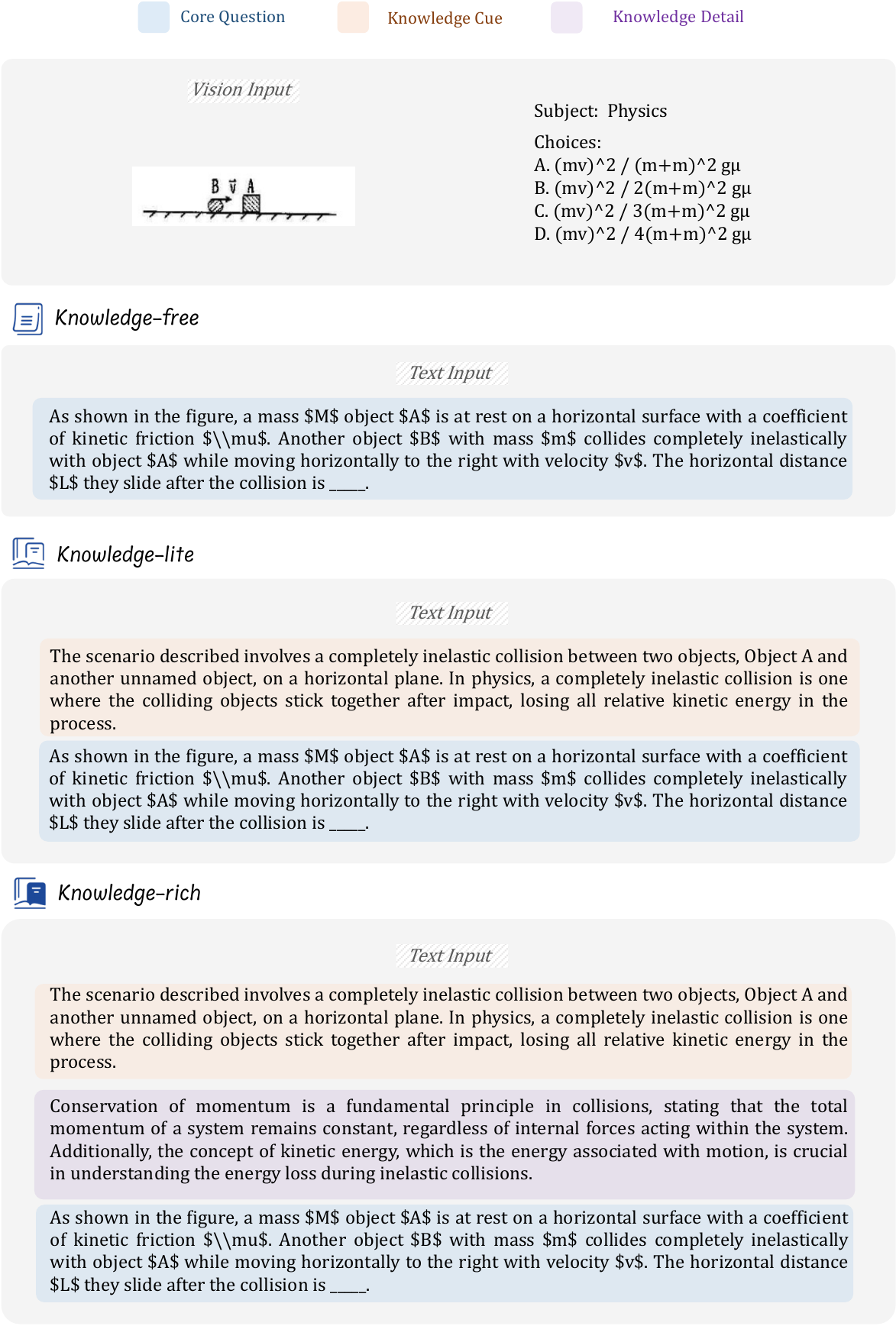}
    \caption{\textbf{Examples of Different Problem Versions in \textsc{SciVerse}.}}
    \label{apfig4}
\end{figure*}

\begin{figure*}[t!]
    \centering
    \includegraphics[width=\linewidth]{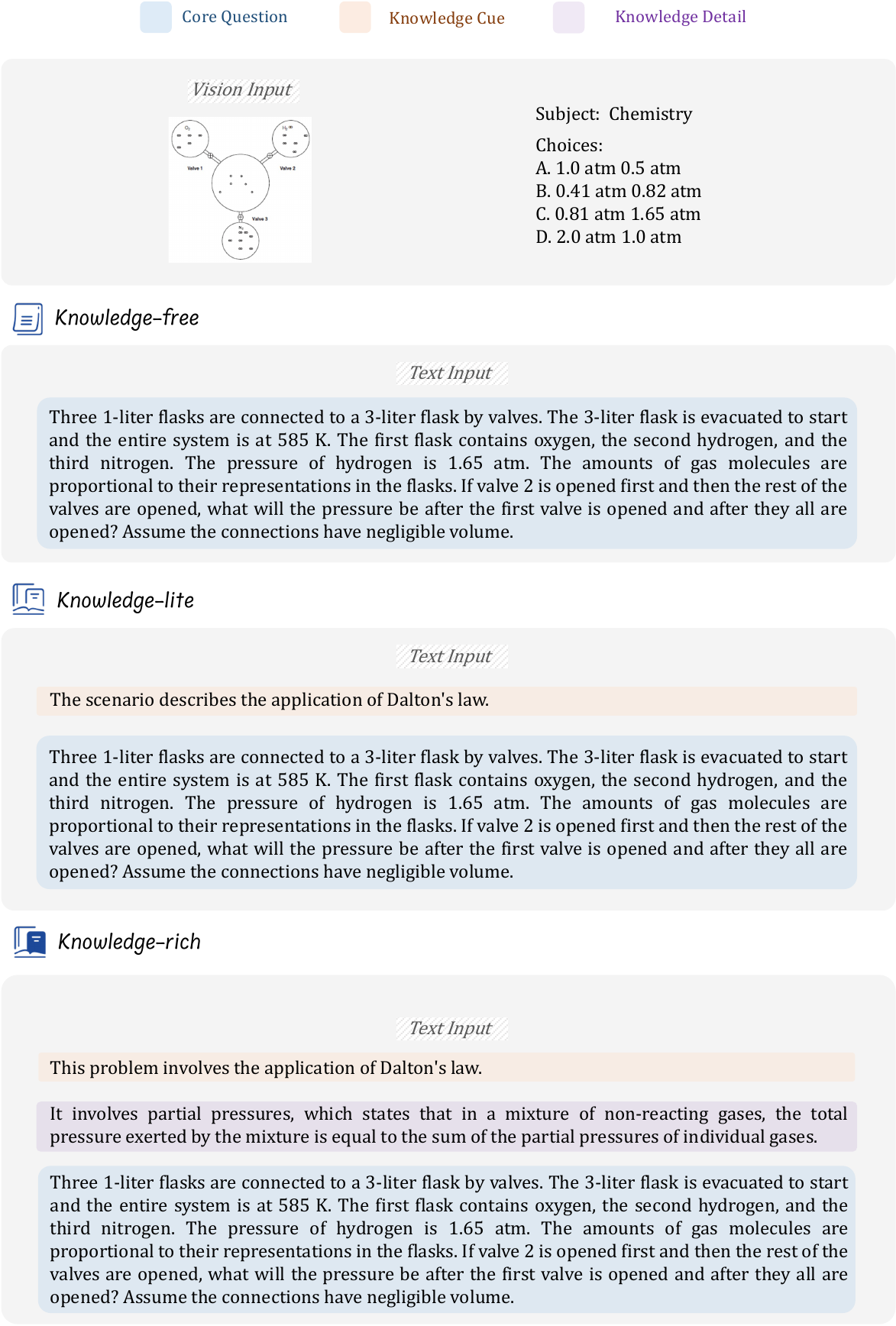}
    \caption{\textbf{Examples of Different Problem Versions in \textsc{SciVerse}.}}
    \label{apfig5}
\end{figure*}

\begin{figure*}[t!]
    \centering
    \includegraphics[width=0.9\linewidth]{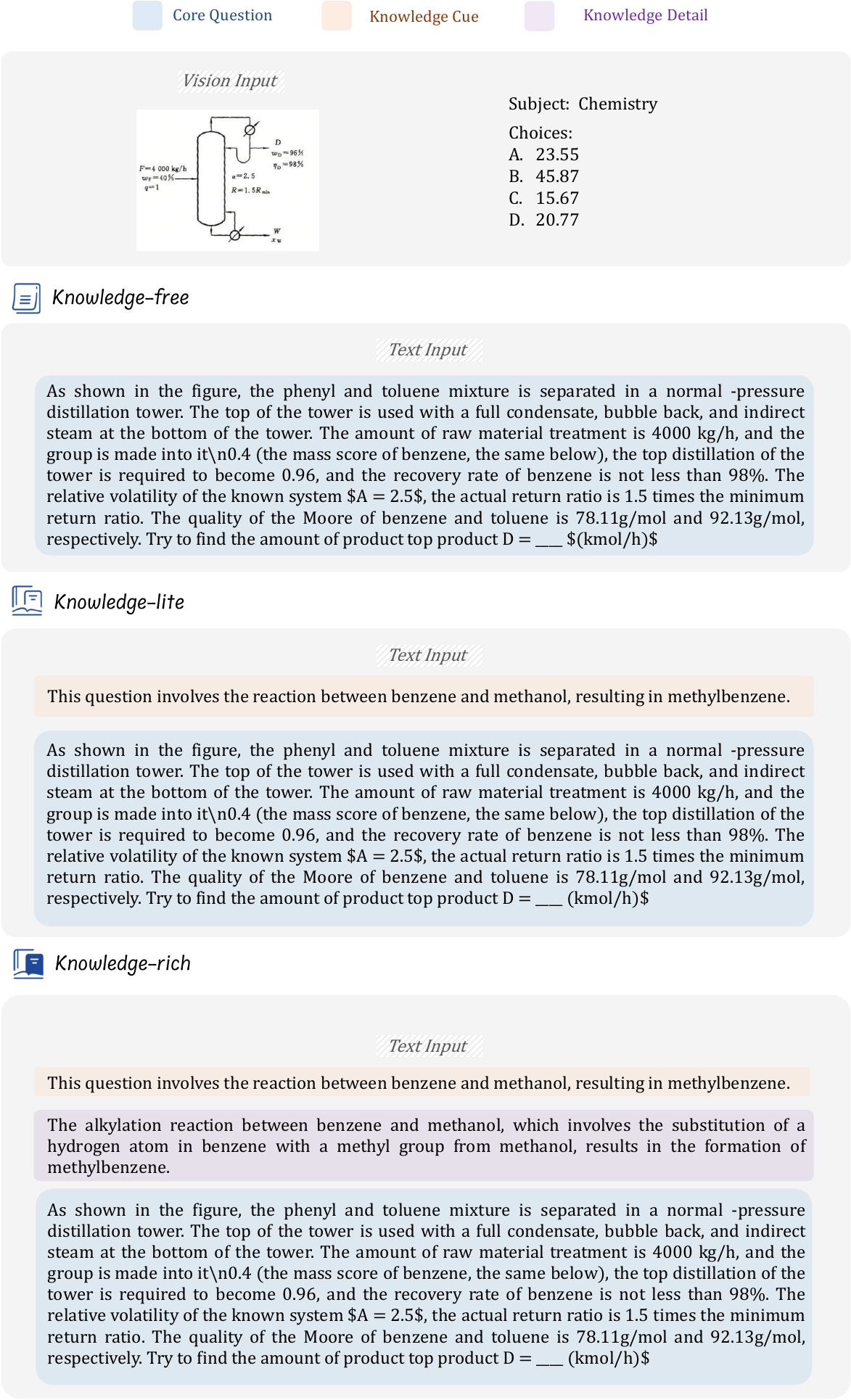}
    \caption{\textbf{Examples of Different Problem Versions in \textsc{SciVerse}.}}
    \label{apfig6}
\end{figure*}

\begin{figure*}[t!]
    \centering
    \includegraphics[width=0.8\linewidth]{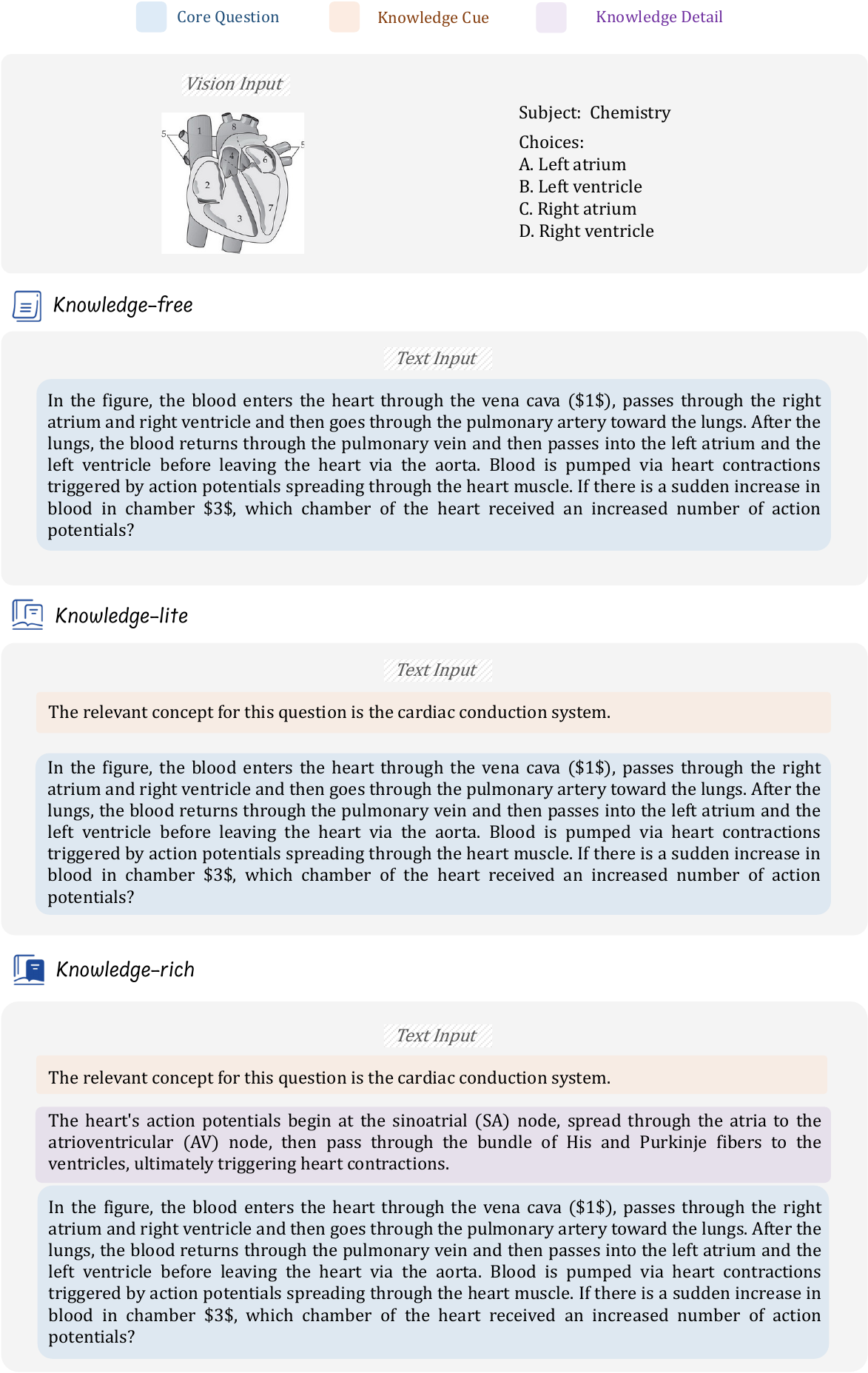}
    \caption{\textbf{Examples of Different Problem Versions in \textsc{SciVerse}.}}
    \label{apfig7}
\end{figure*}

\begin{figure*}[t!]
    \centering
    \includegraphics[width=\linewidth]{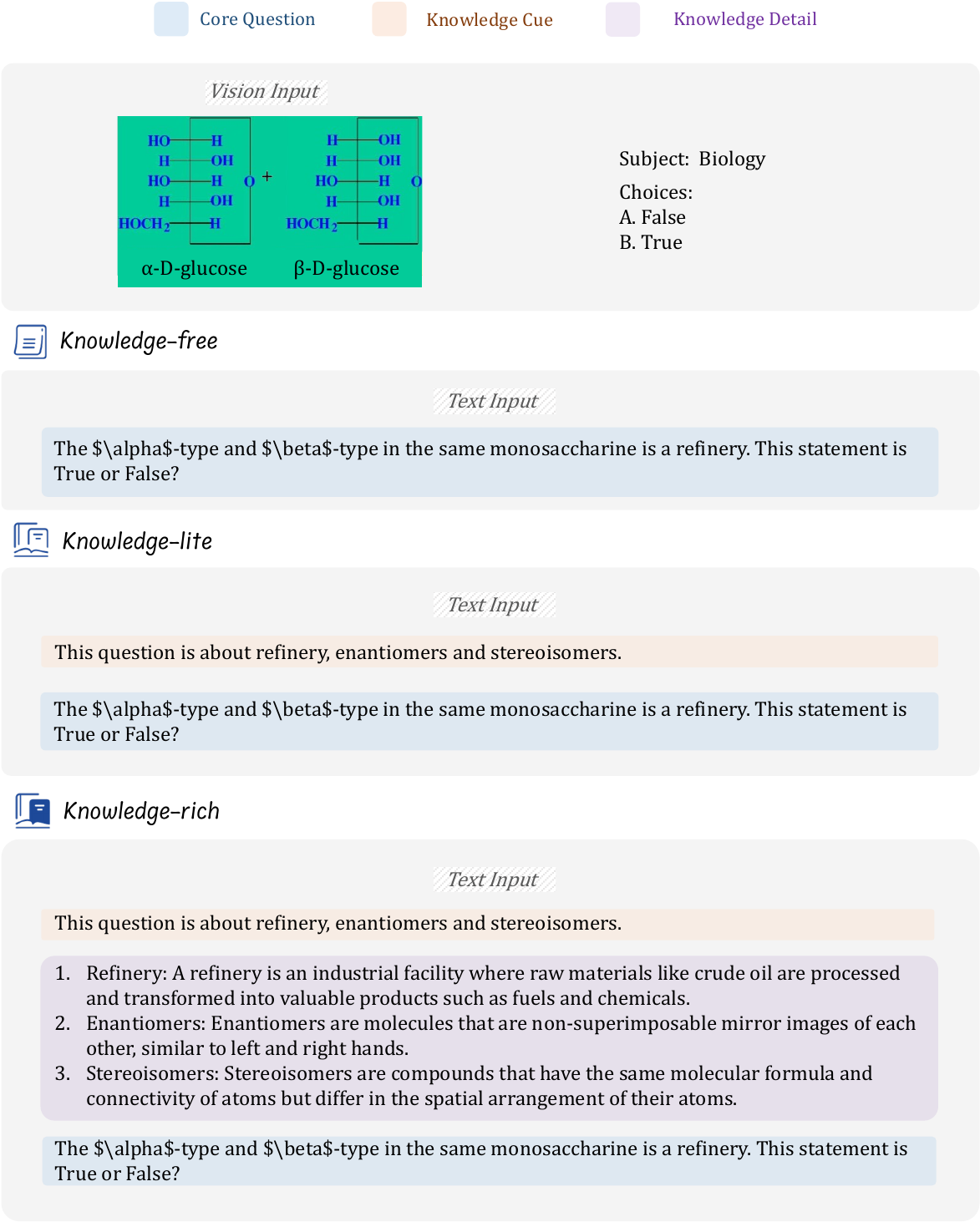}
    \caption{\textbf{Examples of Different Problem Versions in \textsc{SciVerse}.}}
    \label{apfig8}
\end{figure*}

\begin{figure*}[t!]
    \centering
    \includegraphics[width=\linewidth]{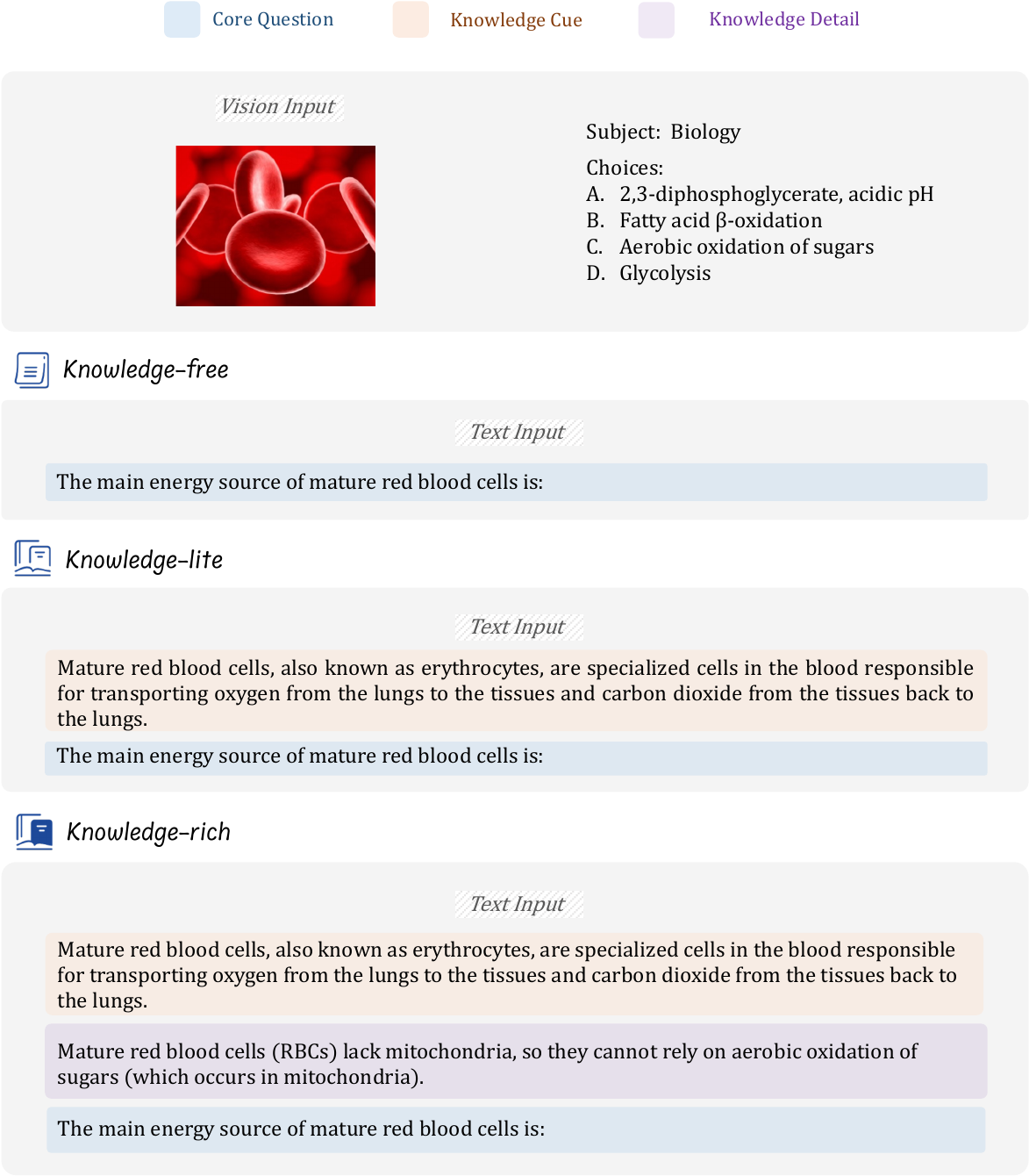}
    \caption{\textbf{Examples of Different Problem Versions in \textsc{SciVerse}.}}
    \label{apfig9}
\end{figure*}

\end{document}